\documentclass[11pt,a4paper]{article}
\usepackage{authblk}
\usepackage[hyperref]{acl2021}

\usepackage{times}
\usepackage{latexsym}
\usepackage{booktabs}

\usepackage{amsmath}
\usepackage{url}
\usepackage{mathdots}
\usepackage{yhmath}
\usepackage{cancel}
\usepackage{color}
\usepackage{siunitx}
\usepackage{array}
\usepackage{multirow}
\usepackage{amssymb}
\usepackage{gensymb}
\usepackage{graphicx}
\usepackage{float}
\usepackage{rotating}
\usepackage{tabularx}
\usepackage{booktabs}
\usepackage{svg}
\usepackage{tablefootnote}
\usepackage{cleveref}
\usepackage{tabularx}
\usepackage{xcolor}
\usepackage{makecell}
\usepackage{enumitem}
\usepackage{color, colortbl}

\definecolor{Gray}{gray}{0.9}

\newcommand{\ours}{\textsc{Pabst}}
\newcommand{\compac}{\textsc{DiscChoice}}
\newcommand{\retrieval}{\textsc{Retrieval}}
\newcommand{\pseudo}{\textsc{Pseudo}}
\newcommand{\personachat}{\textsc{Persona}}
\newcommand{\multitask}{\textsc{MultiTask}}
\newcommand{\transfero}{\textsc{Transfero}}
\newcommand{\commonsense}{\textsc{cs-kb}}

\usepackage{microtype}

\aclfinalcopy 

\title{Unsupervised Retrieval and Rewriting of Short Story Events for More Interesting Chitchat Dialog Responses}
\title{Unsupervised Enrichment of Persona-grounded Dialog\\with Background Stories}

\makeatletter
\renewcommand*{\@fnsymbol}[1]{\ensuremath{\ifcase#1\or \dagger\or \ddagger\or
    \mathsection\or \mathparagraph\or \|\or **\or \dagger\dagger
    \or \ddagger\ddagger \else\@ctrerr\fi}}
\makeatother

\title{Unsupervised Enrichment of Persona-grounded Dialog\\with Background Stories}

\author[$\clubsuit$]{\textbf{Bodhisattwa Prasad Majumder}}
\author[$\clubsuit$]{\textbf{Taylor Berg-Kirkpatrick}}
\author[$\clubsuit$]{\textbf{\qquad \qquad \qquad Julian McAuley}}
\author[$\diamondsuit$]{\textbf{Harsh Jhamtani}}
\affil[$\clubsuit$]{Department of Computer Science and Engineering, UC San Diego \protect\\ \tt \{bmajumde, tberg, jmcauley\}@eng.ucsd.edu}
\affil[$\diamondsuit$]{School of Computer Science, Carnegie Mellon University \protect\\ \tt jharsh@cs.cmu.edu}

\makeatletter
\renewcommand\outauthor{
    \begin{tabular}[t]{>{\centering}p{14cm}} 
    \bf\@author
    \fi
    \end{tabular}}
\makeatother

\date{}

\begin{document}
\maketitle

\begin{abstract}
Humans often refer to personal narratives, life experiences, and events to make a conversation more engaging and rich.
While persona-grounded dialog models are able to generate responses that follow a given persona, they often miss out on stating detailed experiences or events related to a persona,
often leaving conversations shallow and dull.
In this work, we equip dialog models with `background stories' related to a persona by leveraging fictional narratives from existing story datasets (e.g.~ROCStories). 
Since current dialog datasets do not contain such narratives as responses, we perform an unsupervised adaptation of a retrieved story for generating a dialog response using a gradient-based rewriting technique. Our proposed method encourages the generated response to be \textit{fluent} (i.e.,~highly likely) with the dialog history,  \textit{minimally different} from the retrieved story to preserve event ordering and \textit{consistent} with the original persona.
We demonstrate that our method can generate responses that are more diverse, and are rated more engaging and human-like by human evaluators, compared to outputs from existing dialog models.
\end{abstract}

\section{Introduction}

Humans often rely on 
specific incidents and experiences while conversing in social contexts \cite{dunbar1997human}. Responses from existing chitchat dialog agents often lack such specific details. To mitigate this, some prior work has looked into assigning personas to dialog agents \cite{DBLP:conf/acl/KielaWZDUS18, DBLP:conf/emnlp/MajumderJBM20}.  However, persona descriptions are often shallow and limited in scope, and while they lead to improvements response specificity, they still lack the level of detail with which humans share experiences.

\begin{figure}[t]
    \centering
    \includegraphics[width=\linewidth]{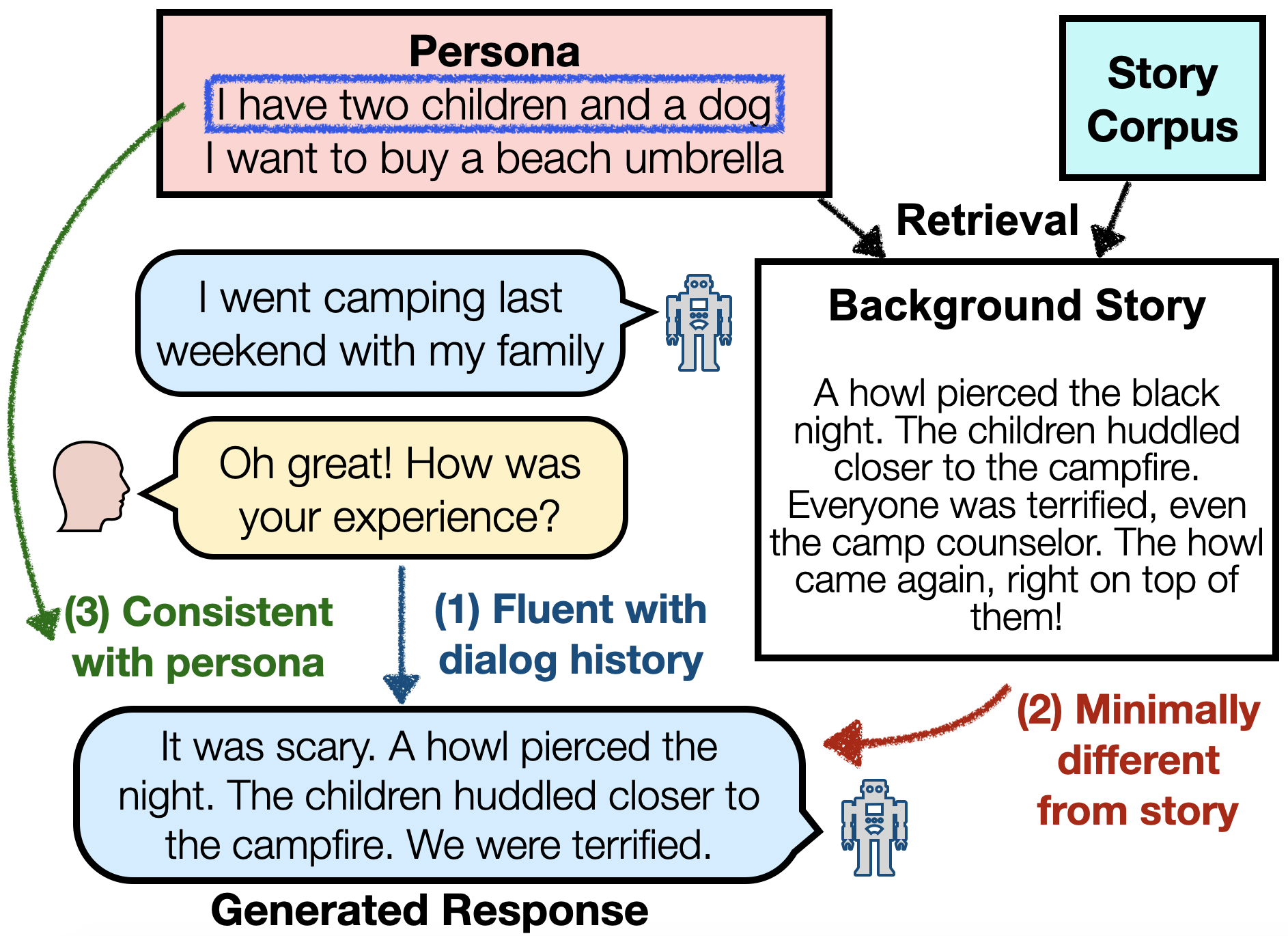}
    \caption{\small We enrich agent personas with `background stories' from an existing corpus. We propose a gradient-based 
    technique
    which encourages the generated response to be fluent with the dialog history, minimally different from the retrieved story, and 
    consistent with the persona. The 
    proposed approach leads to more specific and interesting responses. }
    \label{fig:overview}
    \vspace{-1em}
\end{figure}

In this work, we propose methods to enrich dialog personas with relevant background events using fictional narratives from existing story datasets such as ROCStories \cite{DBLP:journals/corr/MostafazadehCHP16}. For example, for a persona attribute `I have two children and a dog,' we are able to identify a relevant narrative from a story corpus (Figure \ref{fig:overview}).
However, such stories may not directly fit fluently in the dialog context. Thus, retrieved stories should be adapted to construct a response that is fluent and relevant to the context. 
Since existing datasets (such as PersonaChat  \cite{DBLP:conf/acl/KielaWZDUS18}) do not contain responses with such background stories, such adaptation has to be done in an unsupervised fashion with decoders trained to generate responses conditioned only on a dialog history and persona.

To adapt a retrieved narrative incident as a relevant background story, we use a decoding procedure which encourages the generated response to (1) be fluent with the dialog history, (2) be consistent with the original persona, and (3) be minimally different from the retrieved story.
While fluency with dialog context is encouraged directly by the likelihood 
as per the underlying language model
the remaining two constraints are incorporated via iterative updates to the decoder output distributions at inference time. 
Our inference-time decoding method is 
different from the only recent effort by \citet{su2020diversifying} that leverages non-dialog data (forum comments, book snippets) as distant labels to train dialog systems with supervision.
Our contributions can be summarized as follows:
\begin{itemize}[leftmargin=*, itemsep=0.5pt]
    \item We propose a novel approach to enrich dialog agent personas with relevant backstories, relying only on existing story datasets.
    \item We propose to use an unsupervised back-propagation based decoding procedure\footnote{Code can be found at\\ \url{https://github.com/majumderb/pabst}} to adapt the relevant stories such that the resulting response is fluent with the dialog history and consistent with the dialog agent persona. 
    Our method works with a model trained just with dialog data i.e. without access to story corpus at training time.
    \item Our experiments demonstrate that the proposed approach results in much more engaging and specific dialog outputs in a persona-grounded dialog setup. This fills a 
    gap 
    in 
    existing
    dialog models which often lack the capability to generate responses about specific events and experiences relevant to persona attributes.
\end{itemize}

\section{Unsupervised Persona Enrichment with Background Stories}
Given dialog history $h$ and persona $C$ consisting of several (typically 3-5, example shown in Figure \ref{fig:overview}) attributes, our goal is to construct a dialog response $x$. 
Our underlying model is based on the discrete persona attribute choice model from \citet{DBLP:conf/emnlp/MajumderJBM20}. 
To generate a dialog utterance $x$, we first sample a persona attribute $c\sim p(c|h)$ conditioned on the dialog history $h$. 
$x$ is then generated conditioned on the dialog history and the chosen persona attribute. The underlying dialog model's decoder is initialized with a pretrained GPT-2 model, and is fine-tuned on the PersonaChat dataset \cite{DBLP:conf/acl/KielaWZDUS18}.
However, in our current setup, we also have to identify relevant background stories and use them to construct fluent responses at decoding time. Therefore, we propose a different decoding procedure.

To generate a response, we first sample a persona attribute $c\sim p(c|h)$. 
Next
we retrieve stories corresponding to the persona attribute $c$ (Section \ref{sec:retr}). However, the underlying dialog model is trained to generate responses conditioned only on the dialog history and persona. To incorporate the retrieved story in the response, we perform gradient-based inference 
(Section \ref{sec:gradinfer}),
that only assumes a left-to-right language model trained on dialog context and responses, and the story is handled at decoding time in an unsupervised fashion.
We refer to the proposed method as \textbf{\ours{}} (Unsupervised \textbf{P}erson\textbf{A} enrichment with \textbf{B}ackground \textbf{ST}ories).

\subsection{Retrieving Relevant Stories}
\label{sec:retr}

For a persona attribute $c$, we aim to identify relevant stories from a story corpus. 
Toward this goal, we rank the stories using the F1 component of BERT-score \cite{DBLP:conf/iclr/ZhangKWWA20} based retrieval using the persona attribute $c$ as the query and the highest scoring story is chosen.
Note that many of the stories are written in the third person.
For use
as background stories, we must first transform them to 
first–person. 
Following prior work \cite{DBLP:conf/emnlp/BrahmanC20}, we identify the protagonist of such stories as the most frequently occurring character.
Thereafter, we use co-reference resolution \cite{DBLP:conf/emnlp/LeeHLZ17} to identify all 
words or phrases that refer to the protagonist. Finally, all words or phrases so identified are replaced with suitable first person pronouns (e.g.~`his books' to `my books'). 

\subsection{Gradient-based Inference}
\label{sec:gradinfer}
Our underlying dialog model is not trained to condition on a retrieved story, and cannot be directly used to construct a desirable response using $s$.
To tackle this, we consider a decoding strategy which, in addition to fluency with history $h$, encourages response $x$ to follow two soft constraints: (1) be minimally different from story $s$, and (2) be consistent with persona $c$.

First, we generate an initial response based only on the dialog history. Then we perform an iterative procedure which alternates between performing a forward pass on the language model to encourage fluency, and a backward pass which updates the response via back-propagation to respect the two soft constraints. 
However, $x$ is discrete, and cannot be directly updated using gradients from back-propagation. Instead, we maintain and update a soft representation $o$ of $x$, where $o_i$ corresponds to 
the
last hidden state representation for the $i^{th}$ token position, i.e.,~$p(x_i) \sim \operatorname{softmax}(Wo_i/\tau)$, where $\tau$ is the temperature parameter, $W$ is the embedding matrix, and $Wo_i \in \mathcal{R}^V$ ($V$ is the vocabulary size).
Our approach is inspired by recent works that use gradient-based decoding for text generation with soft constraints \cite{DBLP:conf/iclr/DathathriMLHFMY20,DBLP:conf/emnlp/QinSWBHBBC20}.
Next we describe the backward and forward passes of the iterative procedure.

\paragraph{Backward Pass with Soft Constraints}
We define the following 
soft constraints on response $x$: \\
\noindent
(1) \textbf{Divergence from story:} We want to encourage 
$x$ to be \emph{minimally different} from the story $s$. 
Following prior work \cite{DBLP:conf/emnlp/QinSWBHBBC20}, we compute a cross entropy loss (denoted by $\operatorname{cross-entr}$ henceforth) with story $s=\{s_1,\ldots,s_T\}$ tokens as labels and $Wo_1,\ldots,Wo_T$ as the logits.

\noindent
(2) \textbf{Consistency to persona:} We want 
$x$ to be \emph{consistent with persona attribute} $c$. Consider a classifier $q_\phi(o,c)$ which predicts the probability of $x$ (or rather the soft representation $o$ of $x$) entailing $c$. 
The classifier $q_\phi(o,c)$ is a bag-of-words classification head on decoder hidden states $o$, fine-tuned on the Dialogue-NLI dataset \cite{DBLP:conf/acl/WelleckWSC19} to predict whether pairs of persona attributes and responses are entailed or not. 
The objective to maximize can be written as:
\begin{align*}
\resizebox{\linewidth}{!}{%
    $\mathcal{L}(c,s;o) =$ $\lambda_c \log q_\phi(o,c) - \lambda_d  \operatorname{cross-entr}(s,Wo)$%
}
\end{align*}
where $\lambda_c$ and $\lambda_d$ are hyper-parameters.
We update $o$ through back-propagation by computing the gradient $\nabla_{o} \mathcal{L}(c,s;o)$, while keeping the model parameters constant. Let the resulting $o$ after the gradient-based updates be denoted by $o^{b}$. 
\paragraph{Forward Pass to Encourage Fluency}
Next we perform a forward pass of the underlying dialog model
, with the goal of 
regularizing the hidden states towards the unmodified language model values.
On computing the forward pass at the $j^{th}$ token, we mix the 
final hidden states $o^{f}_j$ from the forward pass with $o^{b}_j$ computed in the backward pass, via weighted addition to get the resulting $o_j = \gamma \times o^{f}_j + (1-\gamma) \times o^{b}_j$, where $\gamma\in(0,1)$ is a hyperparameter. The resulting $o_j$ is used for computing the logits at the next time step $j+1$.

We initialize the output response by performing greedy decoding
from the underlying dialog model, 
conditioned on the dialog history and persona attribute. 
Then we iteratively update $o$ by alternate
backward and forward passes. 
We sample the final response $x \sim \operatorname{softmax}(Wo/\tau)$.
In practice, we found that 5 iterations are sufficient to generate good quality outputs. 

\section{Experiments}

\begin{table}[t!]
\small
\centering
\resizebox{\linewidth}{!}{%
\begin{tabular}{@{}l@{\hskip 0.07in}l@{\hskip 0.07in}l@{\hskip 0.07in}c@{\hskip 0.07in}c@{\hskip 0.07in}c@{}}
\toprule
\bf Method & \bf Training & \bf Decoding & \bf D-1 & \bf D-2 & \bf ENTR   \\ \midrule
\rowcolor{Gray}
\multicolumn{6}{l}{\textbf{W/o Story Data}}\\
\transfero{} & \personachat{} & Nucleus & $0.05$ & $0.11$ & $1.21$\\
\compac{} & \personachat{} & Nucleus  & $0.15$  & $0.25$ & $1.25$\\ 
\compac{} & \commonsense{} & Nucleus & $0.87$ & $1.07$ & $2.04$  \\
\midrule
\rowcolor{Gray}
\multicolumn{6}{l}{\textbf{With Story Data}}\\
\compac{} & \pseudo{} & Nucleus & $0.91$ & $2.45$ & $2.89$\\
\compac{} & \multitask{} & Nucleus & $0.99$ & $2.54$ & $2.71$\\
\compac{} & \personachat{} & \retrieval{} & $2.56$ & $9.67$ & $3.86$  \\
\ours{} (Ours) & \personachat{} & Grad. Inf.  & $1.56$ & $3.57$ & $3.21$ \\
\bottomrule
\end{tabular}%
}
\caption{\small Diversity metrics on the PersonaChat test set. D-1/2 is the \% of distinct uni- and bi-grams. ENTR is the geometric mean of n-gram entropy. Grad.~Inf.~is the unsupervised gradient-based decoding as opposed to 
Nucleus sampling \cite{DBLP:conf/iclr/HoltzmanBDFC20}.
}
\label{tab:metrics}
\vspace{-2ex}
\end{table}


\begin{table*}[t!]
\small
\centering
\begin{tabular}{@{}lc@{\hskip 0.05in}c|c@{\hskip 0.05in}c|c@{\hskip 0.05in}c|c@{\hskip 0.05in}c|cc|c@{\hskip 0.05in}c|c@{\hskip 0.05in}c@{}}
\toprule
\bf \ours{}~vs.  & \multicolumn{2}{c|}{\bf \transfero{}} & \multicolumn{2}{c|}{\bf \compac{}} & \multicolumn{2}{c|}{\bf \retrieval{}} & \multicolumn{2}{c|}{\bf \pseudo{}} & \multicolumn{2}{c|}{\bf \multitask{}} & \multicolumn{2}{c|}{\bf w/o DNLI} & \multicolumn{2}{c}{\bf Gold} \\ \midrule
\bf Aspect      & win          & loss   & win          & loss       &  win            & loss           &  win         &  loss         &  win           &  loss      &  win        &  loss    &  win        &  loss      \\ \midrule
Sensible  &  $\mathbf{83.8}$ & $5.6$ & $\mathbf{71.4}$ & $10.1$ & $\mathbf{72.7}$ & $15.7$ & $\mathbf{94.3}$ & $3.1$ & $\mathbf{92.8}$ & $1.5$ & $\mathbf{81.7}$ & $6.8$ & $29.1$ & $\mathbf{33.6}$ \\ 
Engaging &  $\mathbf{96.7}$ & $1.8$ & $\mathbf{82.6}$ & $6.7$ & $\mathbf{87.2}$ & $5.3$ & $\mathbf{94.7}$ & $2.1$ & $\mathbf{94.1}$ & $2.6$ & $\mathbf{70.9}$ & $14.8$ & $\mathbf{70.2}$ & $15.6$ \\  \bottomrule
\end{tabular}
\caption{\small Pairwise comparison (\% win/loss cases, ties not reported) between responses from \ours{} and from other baselines as well as gold. All differences are significant with $p < 0.05$. Cohen's Kappa \cite{cohen1960coefficient} for sensibility and engagement were 0.79 and 0.82 respectively. Complete details of the human evaluations are in Appendix \S B.
}
\label{tab:human1}
\end{table*}

\begin{table*}[t!]
\small
\resizebox{\linewidth}{!}{%
\centering
\begin{minipage}{0.47\textwidth}
\resizebox{0.95\textwidth}{!}{%
\begin{tabular}{@{}l@{}}
\toprule
\begin{tabular}{@{}l@{}}
\textbf{Dialog History} \\
Agent: How was your last trip?\\
User: It was great. I went to Utah and saw the Canyonlands.\\
\midrule
\textbf{Retrieved Story:}\\
I went on a trip. I went to Florida. When we got there, there\\was no air conditioning. I was irritated. It was too hot.\\
\midrule
\midrule
\textbf{\compac{}:}
That's great. You must be adventurous.\\ \midrule
\textbf{\retrieval{}:}
I went on a trip. I went to Florida. When we\\got there, there was no air conditioning. I was irritated.\\It was too hot.\\ \midrule
\textbf{\pseudo{}:}
I have a house in Florida. \\ \midrule
\textbf{\multitask{}:}
I like Florida. I went to the beaches.\\ \midrule
\textbf{\ours{} ($\lambda_d=5$):}
I went on a trip. I went to Florida. I have\\a house there. We do not have air condition. It was hot.\\\midrule
\textbf{\ours{} ($\lambda_d=1$):}
That's great. I frequently go to Florida.\\I have a house there but it gets hot in summer.
\\
\bottomrule
\end{tabular}
\end{tabular}%
}
\end{minipage}
\hfill
\hspace{1em}
\begin{minipage}{0.48\textwidth}
\small
\centering
 \resizebox{\textwidth}{!}{%
\begin{tabular}{@{}l@{}}
\toprule
\begin{tabular}{@{}l@{}}
\textbf{Dialog History} \\
Agent: I work as stunt double. What about you?\\
User: I'm an engineer. I just got back from work. How's yours? \\\midrule
\textbf{Retrieved Story:}\\
I was a working actor. For my new role I was a stunt double. I\\ started working on new stunts. When I jumped off a building\\ yesterday I got hurt. I had to be rushed to the hospital.\\\midrule
\midrule
\textbf{\compac{}:}
That's great. I am looking for a job.\\ \midrule
\textbf{\retrieval{}:}
I was a working actor. For my new role I was a\\stunt double. I started working on new stunts. When I jumped\\off a building yesterday I got hurt. I had to be rushed to the hospital.\\\midrule
\textbf{\pseudo{}:}
I got a new job. I am an actor. \\ \midrule
\textbf{\multitask{}:}
I started a new role. I got hurt.\\ \midrule
\textbf{\ours{} ($\lambda_d=5$):}
I was working as a stunt double. I jumped\\off a building. I got hurt. I had to be rushed to the hospital.\\\midrule
\textbf{\ours{} ($\lambda_d=1$):}
Yesterday while I was working, I jumped off\\a building and I got hurt. I had to be taken to the hospital.\\
\bottomrule
\end{tabular}
\end{tabular}%
}
\end{minipage}%
}
\caption{\small Generations from different models. More examples are in Appendix \S C.}
\label{tab:sample}
\vspace{-3ex}
\end{table*}

We evaluate methods in terms of their capability to generate diverse, fluent and engaging responses. Hyperparameters are noted in Appendix \S A.

\paragraph{Datasets} We experiment with the PersonaChat dialog dataset \cite{DBLP:conf/acl/KielaWZDUS18} 
consisting
of 131,438 utterances for training, 15,602 for validation, and 15,024 for testing. For stories, we use the training split of the ROCStories dataset \cite{DBLP:journals/corr/MostafazadehCHP16}, that consists of 78,529 stories, each typically of 4 to 5 sentences. 

\paragraph{Baselines}
We consider two broad groups of models as baselines: 
\noindent \textbf{(1) \emph{Without access to story corpus}}:
We use finetuned GPT2 (\textbf{\transfero{}}) on PersonaChat, and the discrete persona attribute choice model (\textbf{\compac{}}) from \citet{DBLP:conf/emnlp/MajumderJBM20}. We also consider a version of \compac{} which enriches personas with inferences from a commonsense knowledge base (\textbf{\commonsense{}}).
\noindent \textbf{(2) \emph{Baselines using story corpus}:} To allow \compac{} models to generate story-like responses, we adapt an alternative training regime (\textbf{\pseudo{}}) from \cite{su2020diversifying}, where we randomly replace some of the target dialog responses with retrieved stories---treating them as pseudo labels. 
Finally, we also consider a \textbf{\multitask{}} training setup from \cite{su2020diversifying}, wherein the decoder is trained on PersonaChat as well as with a language modeling objective on ROCStories.
We additionally consider a \textbf{\retrieval{}} baseline that uses the retrieved story verbatim as the dialog response.

\subsection{Automatic Evaluation}
We hypothesize that that the proposed approach to leverage external non-dialog data can increase the diversity of the generated responses. 
Following prior work \cite{DBLP:conf/naacl/LiGBGD16}, we report the percentage of distinct uni-grams and bi-grams (\textbf{D-1} and \textbf{D-2} respectively).
Note that these values do not capture the actual frequency distribution of different word types. Therefore, we also report the geometric mean of entropy values of empirical frequency distributions of n-grams  of words (n $\in \{1,2,3\}$)  \cite{jhamtani2018chess}, denoted by \textbf{ENTR}.

We observe that methods that use story data show much higher diversity compared to methods that do not (Table \ref{tab:metrics}).
Among methods using story data, gradient-based decoding (\ours{}) performs better than \compac{} trained with \pseudo{} or \multitask{}.
Note that just using \retrieval{} outputs as-is leads to even more diverse outputs than \ours{}. However, they are much less sensible with the context, as shown in human evaluations.

\subsection{Human Evaluation}
Since we do not have ground truth story-like responses in the dialog dataset, we perform human evaluation with 150 test examples to investigate if \ours{} generates responses that are 1) \textbf{sensible} with the dialog history and 2) \textbf{engaging}. We hired two Anglophone (Lifetime HIT acceptance \% $>$ 85) annotators for every test sample. The order of the systems present in the interface is randomized. A snapshot of the human evaluation interface is provided in Appendix \S C. All differences in values from human evaluations are significant with $p < 0.05$ from bootstrap tests on 1000 subsets of size 50. Cohen's Kappa \cite{cohen1960coefficient} to measure inter-annotator agreement for sensibility and engagement were 0.79 and 0.82 respectively.

From the results (shown in Table \ref{tab:sample}), we note that in comparison to responses from  baselines, responses from \ours{}~are more engaging and more sensible with respect to the dialog history. 
We further make following observations. 
Firstly, using the gradient-based decoding approach with retrieved stories (\ours{}) works significantly better than using distant supervision with stories data (\pseudo{}~and \multitask{}).
Secondly, background stories provide sufficient detail for an engaging conversation compared to \compac{}~which expands persona attributes using commonsense knowledge \cite{DBLP:conf/emnlp/MajumderJBM20}. 
Finally, we also observe that \ours{}~performs worse when we do not use the consistency constraint (w/o DNLI). 

\paragraph{Choice of $\lambda_d$} We also experiment with different values of the weight for the divergence term ($\lambda_d$) in $\mathcal{L}$: High ($\lambda_d=5$), Moderate ($\lambda_d=1$), and Low ($\lambda_d=0.05$). We consider 100 samples for this experiment. We attribute a high $\lambda_d$ to responses strictly copying the story. We find that \ours{}~(moderate $\lambda_d$) wins wins 81.2\% and 69.1\% cases against \ours{}~(high $\lambda_d$) on `sensible' and `engaging' response criteria respectively. Similarly, \ours{}~(moderate $\lambda_d$) wins 93.2\% and 84.7\% cases against \ours{}~(low $\lambda_d$) in terms of sensibility and engagement respectively. 


\paragraph{Qualitative Analysis}
Table \ref{tab:sample} shows responses generated by different baselines. We observe that
\ours{}
is able to follow the retrieved story (same as output from \retrieval{}) while modifying the response to be conversation-like and sensible with dialog history. 
Responses from other baselines remain verbose or incoherent. Mirroring the human evaluation, we observe that choosing a higher $\lambda_d$ makes the model to almost repeat the retrieved story but a lower value smooths the output to make it more sensible with the ongoing dialog.




\section{Related Work}

A desired impact of the proposed approach is increase in diversity of the generated responses. To tackle the issue of diversity in dialog model outputs, prior work has focused on decoding strategies such as diversity-promoting sampling \cite{DBLP:conf/iclr/HoltzmanBDFC20}; training strategies such as discouraging undesirable responses via unlikelihood training \cite{DBLP:conf/acl/LiRKWBCW20}; model changes such as using stochastic variables \cite{DBLP:conf/aaai/SerbanSLCPCB17}; and using external data such as forum data \cite{su2020diversifying} or external knowledge bases \cite{DBLP:conf/emnlp/MajumderJBM20}. 
In contrast to these, our proposed method
generates responses with background stories using a gradient-based decoding approach.

One of the steps in our proposed approach is to retrieve relevant stories from an external corpus.
Prior work has explored using retrieval of similar dialog instances as an initial step in improving response diversity and other human-like desiderata 
in dialog \cite{roller2020recipes,DBLP:conf/emnlp/WestonDM18}. 
Distant supervision by using retrieved text snippets as pseudo responses has been explored in prior work \cite{su2020diversifying,roller2020recipes}.
We use an external data source to improve dialog responses, a theme shared with some efforts in other tasks such as machine translation \cite{khandelwal2020nearest}.
The use of narrative text in dialog has been explored in prior work, mostly as a `script' or template for conversation \cite{xuenhancing,DBLP:conf/acl/ZhuSDNZ20}. We adapted a BERT-based retrieval method \cite{DBLP:conf/iclr/ZhangKWWA20} in our case to retrieve relevant story given dialog context and use retrieved story in the decoding phase. 

Gradient-based for text generation with soft constraints has been explored in prior work \cite{DBLP:conf/iclr/DathathriMLHFMY20,DBLP:conf/emnlp/QinSWBHBBC20}. \citet{DBLP:conf/aaai/SongZH020} focused on generating response which are consistent to given persona.
Differently, we use a gradient-based decoding to generate a dialog response while honoring constraints such as consistency to persona and similarity to retrieved story.

\section{Conclusion}
We propose a method to enrich persona-grounded dialog with background stories 
at inference time only using 
an
existing corpus of non-conversational narratives---opening up new ways to generate enriched and engaging responses. One of the limitations of \ours{}~is the assumption of the need of a background story at every turn. As future work, we can include a decision step to decide if we need to incorporate a background story or not, given the dialog history. We can further explore ways to use retrieved stories over multiple turns instead of a single turn.

\section*{Acknowledgements} We thank anonymous reviewers for providing valuable feedback. BPM is partly supported by a Qualcomm Innovation Fellowship and NSF Award \#1750063. Findings and observations are of the authors only and do not necessarily reflect the views of the funding agencies.

\section*{Impact Statement}
In this work, we discuss ways to make a dialog system to generate more engaging responses. 
Since we use a finetuned version of a pretrained generative model, we inherit the general risk of generating biased or toxic language, which should be carefully filtered. Furthermore, the generations may incorporate biases that are already present in the dialog dataset and story dataset due to crowd-sourced data collection. Hence, we cautiously advise any developer who wishes to use a different story dataset for the background stories to be aware of the biases present in the dataset.
Finally, we also note that experiments in this paper are limited only to English language.

\bibliography{acl2021.bib}
\bibliographystyle{acl_natbib}

\clearpage

\appendix


\section{Implementation Details}
We obtain the PersonaChat dataset from ParlAI repository\footnote{\url{http://parl.ai/downloads/personachat/personachat.tgz}}. ROCStories dataset is obtained from the repository of original release\footnote{\url{https://www.cs.rochester.edu/nlp/rocstories/}}.
We adapted codes from original PPLM \cite{DBLP:conf/iclr/DathathriMLHFMY20} repository\footnote{\url{https://github.com/uber-research/PPLM}} and modified them for our own objective function. 
\paragraph{Network architecture}
For the generator network, we use GPT2 (Transformer with 12 layers, 768 hidden size, 12 heads--- \texttt{gpt2-small}\footnote{\url{https://github.com/huggingface/transfer-learning-conv-ai}}) following the state-of-the-art model \cite{DBLP:journals/corr/abs-1901-08149} from Conv-AI2 competition.
The decoder has total of 124 Million parameters. We used the pre-trained decoder model obtained from \cite{DBLP:conf/emnlp/MajumderJBM20}.

\paragraph{Hyperparameters}
\ours{}~does not require any training since we perform gradient-based decoding at the inference time. For our best method, in objective function $\mathcal{L}$, we use $\lambda_d$ as 1 and $\lambda_c$ as 1. We keep generation length to be 100 to encourage longer generations. We train the consistency classifier using code from PPLM repository\footnote{\url{https://github.com/uber-research/PPLM/blob/master/run_pplm_discrim_train.py}}. The weight $\gamma$ for mixing forward and backward passes was set to 0.45.
For \pseudo{}, we substitute a random 30\% of the original target responses with retrieved stories. 

\section{Human Evaluations Setup}
\begin{figure*}[t]
    \centering
    \includegraphics[width=\linewidth]{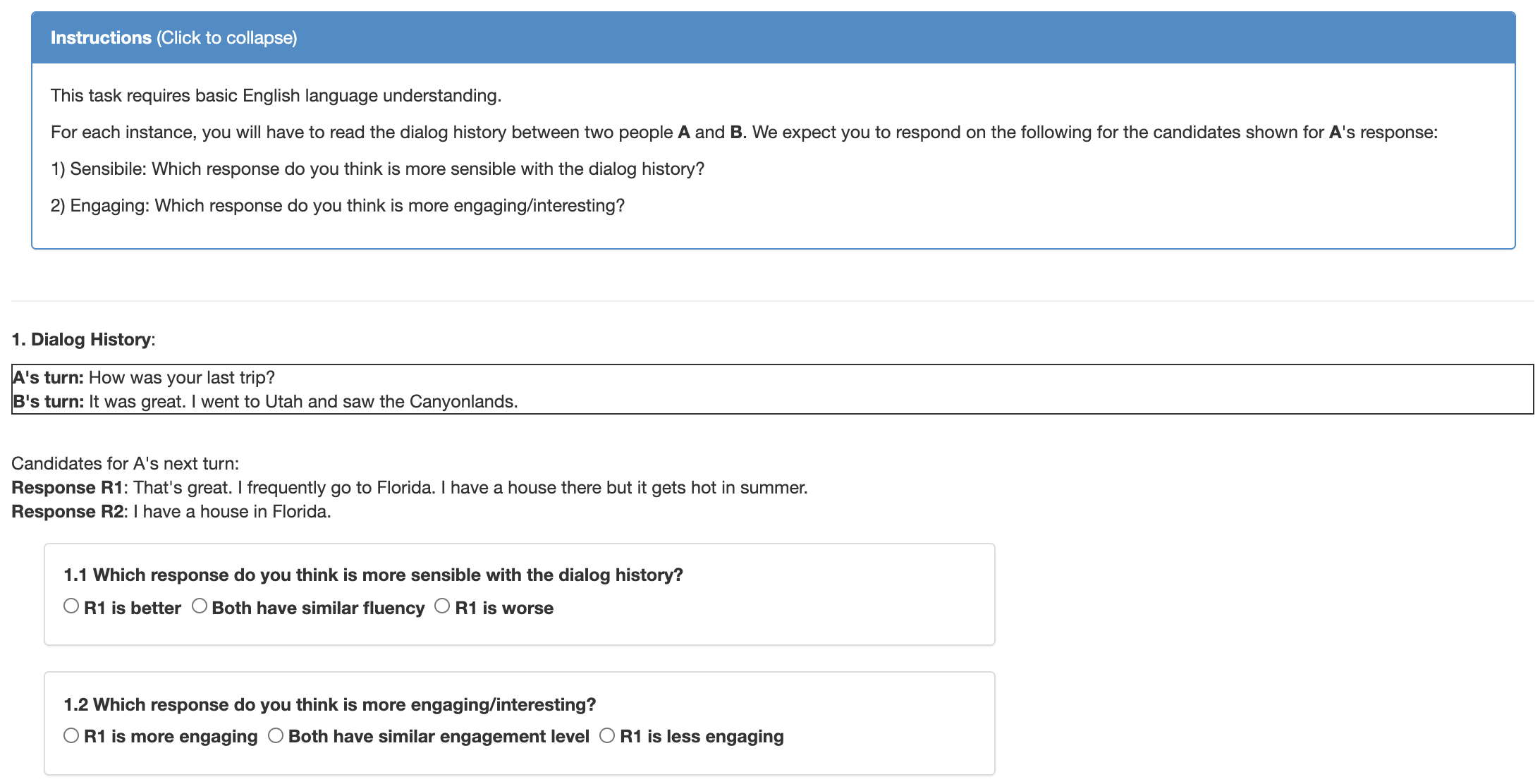}
    \caption{Human evaluation setup for pairwise comparison between \ours{}~and another baseline}
    \label{fig:human_eval1}
\end{figure*}

\Cref{fig:human_eval1} shows a sample question for the pairwise comparison between response generated by \ours{} and a baseline for sensibility and engagingness. A similar setup was used to measure performance between \ours{} variants with different $\lambda_d$ values ($0.5, 1, 5$).

\section{Qualitative Examples}
\Cref{tab:sample2,tab:sample3} show generations from \ours{} and baselines for sampled text examples.


\begin{table}[h!]
\small
\begin{tabular}{@{}l@{}}
\toprule
\begin{tabular}{@{}l@{}}
\textbf{Dialog History} \\
Agent: I volunteer in shelters.\\ 
User: It is a noble act.\\\midrule
\midrule
\textbf{\compac{}:}
Thank you. I like helping people.\\ \midrule
\textbf{\retrieval{}:}
I was at my house bored. I couldn't figure out\\what to do. I looked on TV for answers. I saw a commercial\\about homeless people. So I decided to volunteer at a\\homeless shelter.\\ \midrule
\textbf{\pseudo{}:}
Last week, I went to the local animal shelter on\\the pet adaptation day.
\\ \midrule
\textbf{\multitask{}:}
Thank you! What about you?\\ \midrule
\textbf{\ours{} ($\lambda_d=5.0$):}
I was bored. I wanted to help people. I\\looked on TV about homeless people. So I decided to\\volunteer at a homeless shelter.\\\midrule
\textbf{\ours{} ($\lambda_d=1.0$):}
Thank you. I was inspired by a TV\\commercial and I volunteered at the homeless shelter.
\\
\bottomrule
\end{tabular}
\end{tabular}
\caption{Generations from different models}
\label{tab:sample2}
\end{table}

\begin{table}[t!]
\small
\begin{tabular}{@{}l@{}}
\toprule
\begin{tabular}{@{}l@{}}
\textbf{Dialog History} \\
Agent: I enjoy American sports. What sports do you like? \\ 
User: I love playing volleyball. I was in the university team.\\\midrule
\midrule
\textbf{\compac{}:}
That's great. I like adventure sports.\\ \midrule
\textbf{\retrieval{}:}
I always thought golf was for old people.\\One day, my dad invited me to play golf with him.\\I reluctantly accepted his invitation. On the golf range, I\\saw lots of people of my age. I realized golf was a sport for\\everyone to enjoy regardless of age.\\ \midrule
\textbf{\pseudo{}:}
Last week, I went to play golf. My dad invited\\me. I was reluctant.
\\ \midrule
\textbf{\multitask{}:}
That's great! I play volleyball during\\weekend.\\ \midrule
\textbf{\ours{} ($\lambda_d=5.0$):}
I like playing golf. I always thought it\\is for old people. Recently I went to a golf range. But I saw\\lots of people of my age. I realized golf is a\\sports for everyone.\\\midrule
\textbf{\ours{} ($\lambda_d=1.0$):}
That's great. I like playing golf. I\\always thought it is for old people. Recently I went\\to a golf range. But I saw lots of people of my age. I\\realized golf is a sports for everyone.
\\
\bottomrule
\end{tabular}
\end{tabular}
\caption{Generations from different models}
\label{tab:sample3}
\end{table}

\end{document}